\documentclass{article}

\usepackage{microtype}
\usepackage{graphicx}
\usepackage{subfigure}
\usepackage{booktabs} 

\usepackage{hyperref}

\newcommand{\comm}[1]{}

\usepackage[accepted]{icml2025}

\usepackage{amsmath}
\usepackage{amssymb}
\usepackage{mathtools}
\usepackage{amsthm}

\usepackage[capitalize,noabbrev]{cleveref}

\theoremstyle{plain}

\theoremstyle{definition}

\theoremstyle{remark}

\usepackage[textsize=tiny]{todonotes}

\icmltitlerunning{Lyapunov Learning at the Onset of Chaos}

\begin{document}

\twocolumn[
\icmltitle{Lyapunov Learning at the Onset of Chaos}




\begin{icmlauthorlist}
\icmlauthor{Matteo Benati}{yyy}
\icmlauthor{Alessandro Londei}{xxx}
\icmlauthor{Denise Lanzieri}{xxx}
\icmlauthor{Vittorio Loreto}{xxx,zzz,ppp}
\end{icmlauthorlist}
\icmlaffiliation{yyy}{Department of Computer, Automatic and Management Engineering.
  Sapienza University, Via Ariosto 25, Rome, Italy \\}
\icmlaffiliation{xxx}{Sony Computer Science Laboratories - Rome. Joint Initiative CREF-SONY, Centro Ricerche Enrico Fermi. Via Panisperna 89/A, 00184, Rome, Italy \\}
\icmlaffiliation{zzz}{Sapienza University of Rome, Physics Department.Piazzale A. Moro, 2, 00185, Rome, Italy, \\}
\icmlaffiliation{ppp}{Complexity Science Hub \\
Josefstädter Strasse 39, A 1080, Vienna, Austria}
\icmlcorrespondingauthor{Matteo Benati}{matteo.benati@uniroma1.itu}

\icmlkeywords{Machine Learning, ICML}

\vskip 0.3in
]



\printAffiliationsAndNotice{}  

%

\author{%
  David S.~Hippocampus\thanks{Use footnote for providing further information
    about author (webpage, alternative address)---\emph{not} for acknowledging
    funding agencies.} \\
  Department of Computer Science\\
  Cranberry-Lemon University\\
  Pittsburgh, PA 15213 \\
  \texttt{hippo@cs.cranberry-lemon.edu} \\
}

\begin{abstract}
Handling regime shifts and non-stationary time series in deep learning systems presents a significant challenge. In the case of online learning, when new information is introduced, it can disrupt previously stored data and alter the model's overall paradigm, especially with non-stationary data sources. Therefore, it is crucial for neural systems to quickly adapt to new paradigms while preserving essential past knowledge relevant to the overall problem.
In this paper, we propose a novel training algorithm for neural networks called \textit{Lyapunov Learning}. This approach leverages the properties of nonlinear chaotic dynamical systems to prepare the model for potential regime shifts. Drawing inspiration from Stuart Kauffman's Adjacent Possible theory, we leverage local unexplored regions of the solution space to enable flexible adaptation. The neural network is designed to operate at the edge of chaos, where the maximum Lyapunov exponent, indicative of a system's sensitivity to small perturbations, evolves around zero over time.
 Our approach demonstrates effective and significant improvements in experiments involving regime shifts in non-stationary systems. In particular, we train a neural network to deal with an abrupt change in Lorenz's chaotic system parameters. The neural network equipped with Lyapunov learning significantly outperforms the regular training, increasing the loss ratio by about $96\%$.
\end{abstract}
\section{Introduction}
Integrating new data in machine learning systems is a vital area of study, especially regarding the issues of catastrophic forgetting and continual learning \citep{mccloskey1989catastrophic, ratcliff1990connectionist, hadsell2020embracing}. 
A significant focus in this field is integrating new data after a neural network has undergone its initial training without disrupting or overwriting previously learned information \citep{10444954}. This problem is exacerbated when the new data has statistical properties different from the previously stored information. In such scenarios, the system must quickly adjust its parameters to accurately reflect the latest information while preserving earlier knowledge\citep{NEURIPS2022_4054556f,dreaminglearning2024}. \newline
As Stuart Kauffman points out \citep{Kauffman2000-KAUI-4,Tria2014}, innovations occur when existing information is recombined according to a system's evolutionary rules, expanding the current element set. 
This process is encapsulated in the concept of the \textbf{Adjacent Possible}, which defines the system’s evolving capacity to explore new possibilities through slight modifications of known elements.
Typically, machine learning lacks tools that allow neural networks to effectively explore novel information to better adapt to potential incoming regime shifts.
In this paper, we introduce a novel method to assess and manage the Lyapunov exponent spectrum of an artificial neural system. Lyapunov exponents quantify the rate at which nearby trajectories in a dynamical system diverge or converge, facilitating the network to explore and detect specific directions in phase space. These directions help balance divergence and convergence along the trajectory set by the training data. This can potentially prepare the network for eventual upcoming regime shifts and enhance the network's ability to adapt to novelties. This balance, known as the onset of chaos  \citep{10.1016/0167-2789(90)90064-V, PhysRevLett.84.5991, DBLP:journals/corr/abs-2107-09437} is achieved by controlling the Lyapunov exponents during training to identify suitable divergent directions while preserving the system’s overall dissipative state.  
Our experiments first demonstrate the controlled chaoticity of a neural network by imposing suitable Lyapunov spectrum's target values and obtaining well-defined chaotic attractors spontaneously generated. Then, we apply a Lyapunov-based regularizer to enhance the network behavior in case of a regime shift given by a phase transition of Lorenz's chaotic system data.

\section{Lyapunov Learning's framework}
\paragraph{Lyapunov exponents and chaotic dynamics}
Chaotic systems are dynamical systems characterized by unpredictable behavior, non-repeating patterns, and sensitive dependence on initial conditions. Despite this seemingly random behavior, the system's trajectories remain confined to a specific region in its phase space, indicating an underlying order or structure. The long-term behavior of such systems, i.e., the set of states toward which the dynamical system evolves, is described by chaotic attractors. To identify a chaotic attractor, we analyze the system's Lyapunov exponents.
 For a multidimensional system described by $\frac{d\mathbf{x}}{dt} = \mathbf{f}(\mathbf{x})$, the Lyapunov exponents $\{\lambda_i\}$ quantify how small perturbations $\delta \mathbf{x}$ evolve over time. In this context, the exponents are determined from the eigenvalues of the matrix product \citep{strogatz:2000}:
\begin{equation}
\Lambda = \lim_{T \to \infty} \frac{1}{T} \ln\left|\prod_{t=0}^{T}\mathbf{J}(\mathbf{x}_t)\right|,
\label{Lambdamatrix}
\end{equation}
where $\mathbf{J}(\mathbf{x}_t)$ is the Jacobian matrix of the system at time $t$. These exponents describe the rate of growth or decay of perturbations in the directions defined by the eigenvectors. Specifically, a positive exponent indicates exponential divergence (instability), a negative exponent suggests convergence to a fixed point, and a zero-valued exponent implies periodic behavior. The number of Lyapunov exponents corresponds to the number of dimensions in the system.
There are two key conditions involving Lyapunov exponents that are necessary to confirm the existence of a chaotic attractor:\newline
    \textbf{1} \textbf{At least one positive Lyapunov exponent:} This indicates that nearby trajectories diverge exponentially in the direction of the corresponding eigenvector, ensuring chaotic behavior. \newline
    \textbf{2} \textbf{A negative sum of all Lyapunov exponents:} This condition ensures that the system's trajectories remain bounded within a finite region, reflecting the dissipative nature of chaotic attractors.
\paragraph{Lyapunov Learning's algorithm}
If we consider the neural network's structure as a dynamical system, $\mathbf{\tilde{x}}_{t+1} = \mathbf{F}(\mathbf{x}_t, \mathbf{w})$, where $\mathbf{x}_t$ represents the data and $\mathbf{w}$ denotes the network's weights, we can estimate Lyapunov exponents. We first generate a sequence starting from real data and evolving through recurrent applications of the network $\mathbf{F}(\mathbf{x}_t, \mathbf{w})$. Then, we compute the Jacobian matrix at each step of the generated sequence. At the end of this process, we apply \autoref{Lambdamatrix} for estimating the Lyapunov exponents. These exponents are calculated over a finite time $T$ (\autoref{Lambdamatrix}), using approximation techniques \citep{Abarbanel1997-tq}. A key feature of this algorithm is its differentiability with respect to the network's weights, thus making possible to integrate Lyapunov exponents directly into the network's loss function. Specifically, to bring the network closer to the onset of chaos and encourage exploration of its space of possibility, we add an additional term, $\mathcal{L}_{\text{Lyapunov}}$ to the original loss. The overall loss function is given by:
\begin{equation}
 \mathcal{L}(\mathbf{x}_t, \mathbf{\tilde{x}}_t) = \mathcal{L}_{\text{data}} + \alpha \cdot \mathcal{L}_{\text{Lyapunov}}
\label{Loss1}
\end{equation}
where $\mathcal{L}_{\text{data}}$ is the task-specific error, and $\alpha$ is the weight of the Lyapunov-based term $\mathcal{L}_{\text{Lyapunov}}$.
\section{Results}
\paragraph{Assessment and control of chaoticity in neural networks}

\begin{figure}[h!]
    \centering
    \begin{minipage}[h!]{0.30\textwidth}
        \centering
        \includegraphics[width=\textwidth]{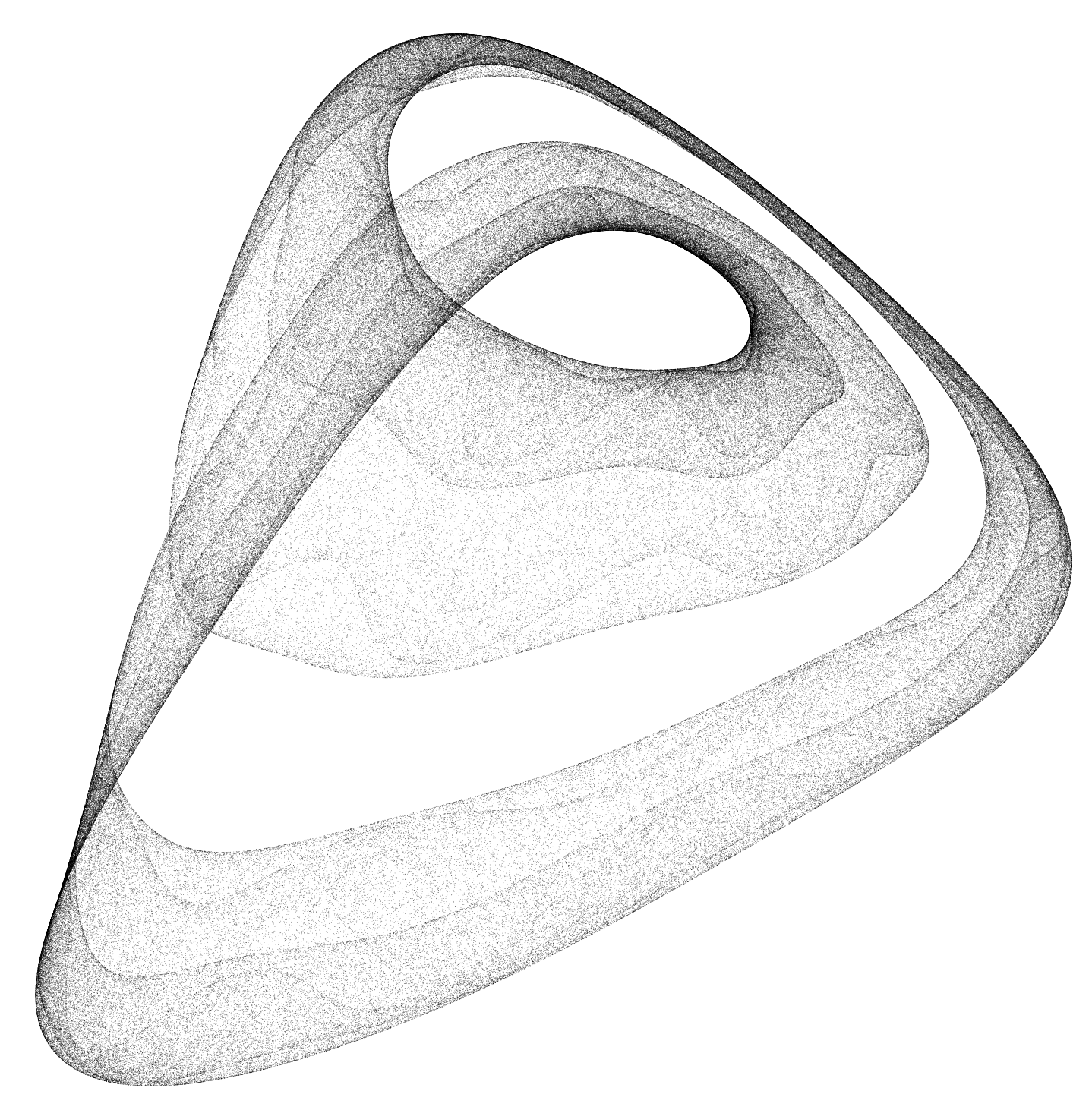}
        \label{fig:attractor1}
    \end{minipage}
    \hfill
        \begin{minipage}[h!]{0.30\textwidth}
        \centering
        \includegraphics[width=\textwidth]{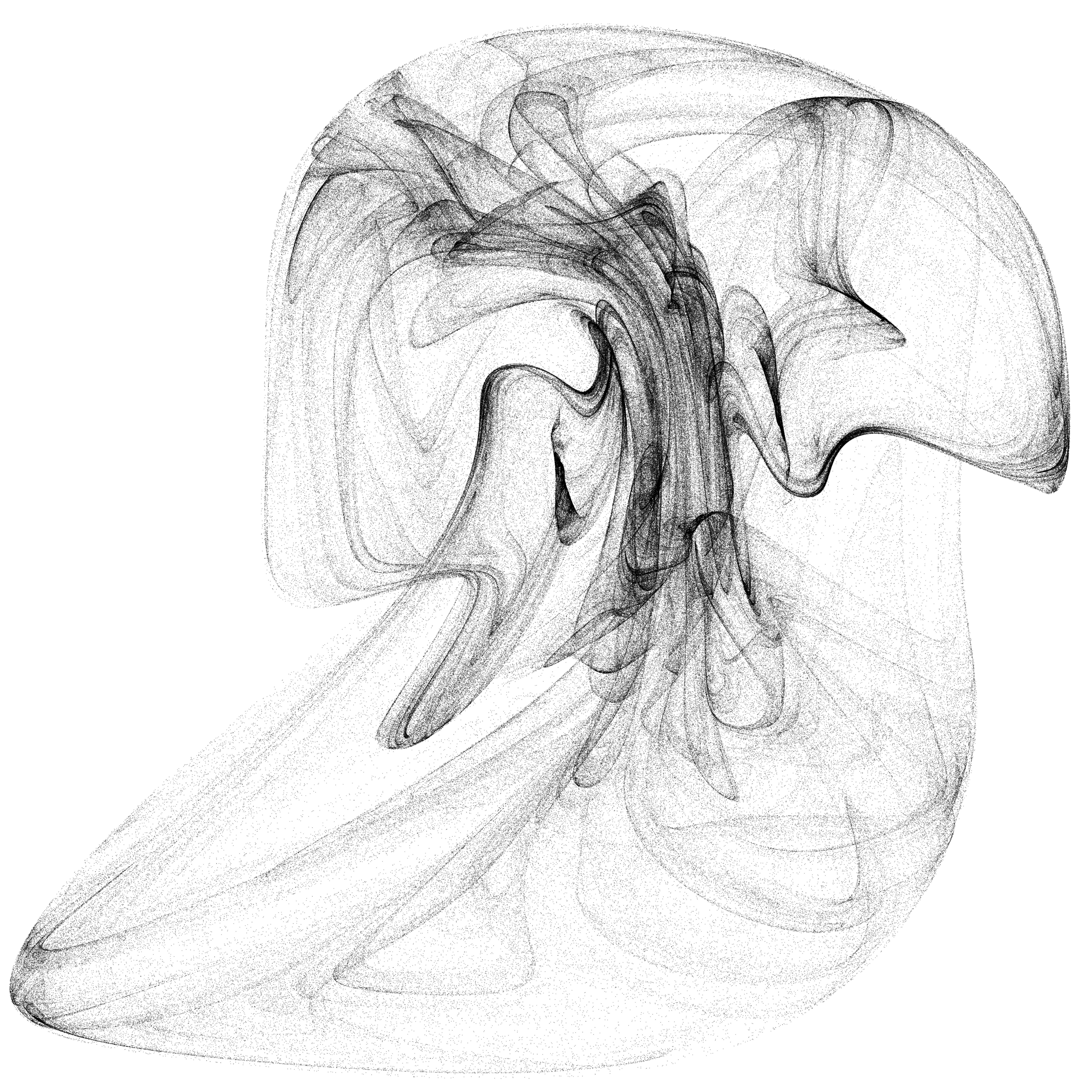}
        \label{fig:figure3}
    \end{minipage}
    \hfill
    \begin{minipage}[h!]{0.30\textwidth}
        \centering
        \includegraphics[width=\textwidth]{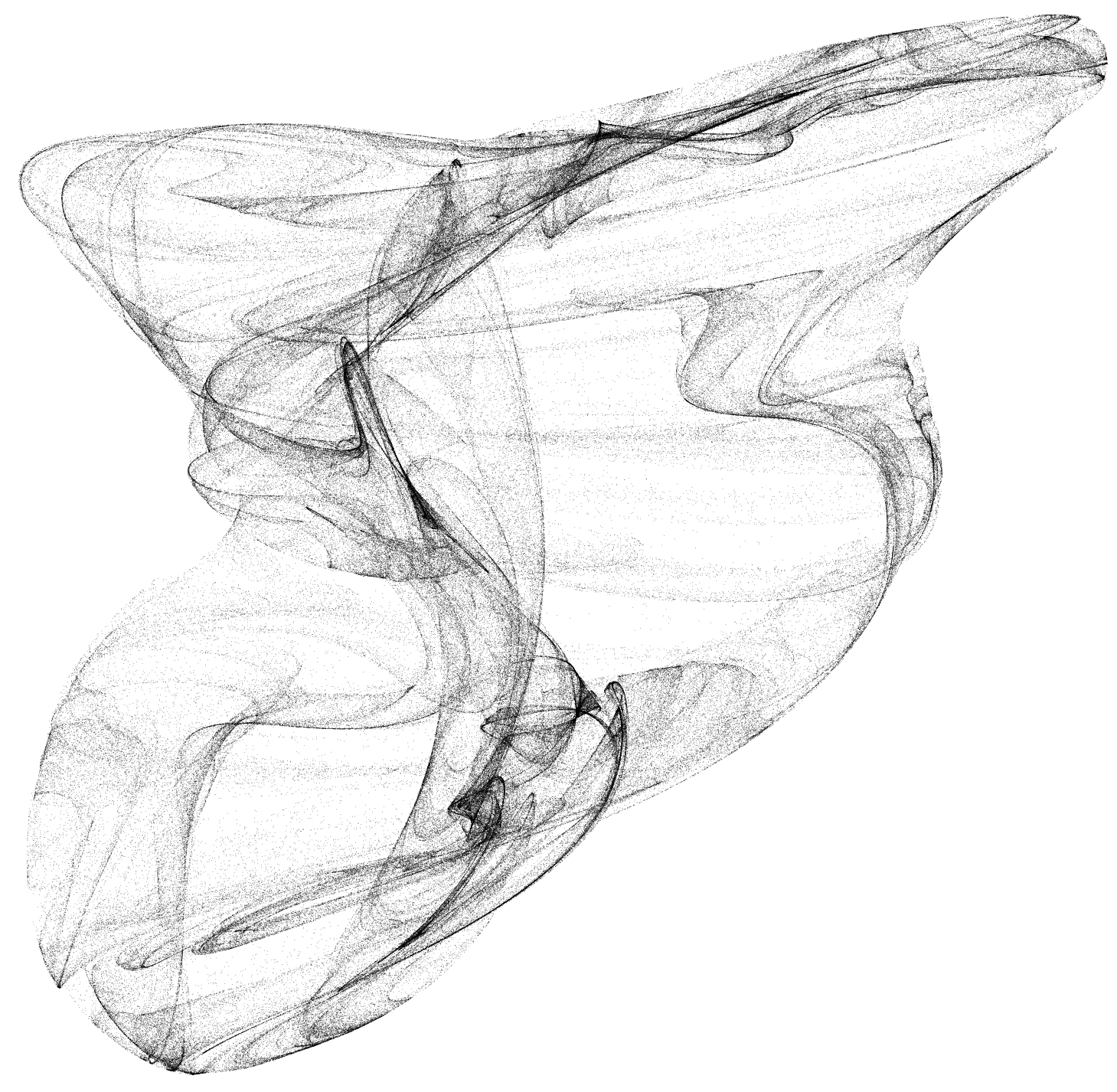}
        \label{fig:figure3}
    \end{minipage}
    \caption{Example of chaotic attractors generated after training, starting from a single initial point after a million iterations. The maximum Lyapunov exponents are 0.104, 0.191, and 0.235, respectively, from top to bottom.}
    \label{fig:attractor}
\end{figure}

Before testing Lyapunov Learning on an application involving regime shifts, we first seek to validate whether our algorithm accurately computes the Lyapunov exponents of sequences generated by a neural network. This step is crucial to ensure we can effectively control and predict the network's behavior by directly influencing these exponents.
To achieve this, we design a network with the specific goal of generating chaotic attractors, where the loss function depends solely on the Lyapunov exponents computed from the sequences produced by the network.
Starting with a single three-dimensional point as input, the network autonomously generates sequences from which the Lyapunov exponents are calculated. These exponents guide the learning process, minimizing the loss function when the network successfully produces sequences that form chaotic attractors.
The network architecture consists of a single hidden layer with 10 neurons. The three plots shown in \autoref{fig:attractor} represent long trajectories of single initial points after a transient phase of 1000 iterations. All trajectories exhibit a positive largest Lyapunov exponent $\lambda$ (condition \textbf{1}), indicating chaotic behavior. As $\lambda$ increases, this chaotic behavior intensifies, reflecting greater sensitivity to initial conditions. All trajectories are also bounded in a specific region of the phase space and the sum of their Lyapunov exponents is strictly negative (condition \textbf{2}).
The successful generation of multiple chaotic attractors, as evidenced by their Lyapunov spectra, confirms the validity of the proposed method.

\paragraph{Lorenz map regime shift}
We now turn to a scenario where the Lyapunov term,  $\mathcal{L}_{\text{Lyapunov}}$ is used as a regularizer to the actual loss $\mathcal{L}_{\text{data}}$, as in \autoref{Loss1}.
A Lorenz map is a mathematical framework used to analyze the dynamics of chaotic systems, especially those exhibiting chaotic behavior. It describes how points in the system evolve through the following equations:
\begin{equation}
\frac{dx}{dt} = \sigma (y - x); \hspace{10pt}
\frac{dy}{dt} = x (\rho - z) - y; \hspace{10pt}
\frac{dz}{dt} = x y - \beta z.
\label{LorenzEquation}
\end{equation}
In this study, we simulate a regime shift in a Lorenz map time series by abruptly changing its parameters halfway through the data. We then use this time series to train two networks: a regular one (hereafter vanilla) and one equipped with Lyapunov Learning. We aim to determine whether the Lyapunov regularizer can improve the network's robustness and flexibility during the regime shift. The two networks are feed-forward networks with 4 layers, 50 neurons each. The network’s input consists of three-dimensional coordinates derived from the data. Once the output, which represents the network's prediction for the next point, is obtained, we compute the network’s $3 \times 3$ Jacobian matrix. We iterate this process multiple times and use QR decomposition to accurately estimate the eigenvalues of the $\Lambda$ matrix \citep{Abarbanel1997-tq}. We focus on the largest Lyapunov exponent, $\lambda$, and do not enforce the dissipativity condition because our vanilla training simulations consistently show negative Lyapunov exponents, which indicates that the condition is naturally satisfied. The total loss function thus becomes:
\begin{equation} 
    \mathcal{L}(x,\hat{x}) = \mathcal{L}_{\text{MSE}}(x,\hat{x}) + \alpha|\lambda|. 
\end{equation}
Here, $\lambda$ is the largest Lyapunov exponent, and $\alpha$ controls the influence of the Lyapunov regularizer. We apply the absolute value of $\lambda$ because, unlike in chaotic attractor generation, we aim to push the network to the edge of chaos without allowing it to become fully chaotic.
We choose the initial and final parameters of our Lorenz maps as follows: initially, we set \( \sigma = 20 \), \( \beta = \frac{8}{3} \), and \( \rho = 28 \), which defines a trajectory that slowly converges. Halfway through the training sequence, we switch to the classic Lorenz chaotic attractor, with parameters \( \sigma = 10 \), \( \beta = \frac{4}{3} \), and \( \rho = 28 \). 
\begin{figure}[h!]
    \centering
    \begin{minipage}[h!]{0.495\textwidth}
        \centering
        \includegraphics[width=\textwidth]{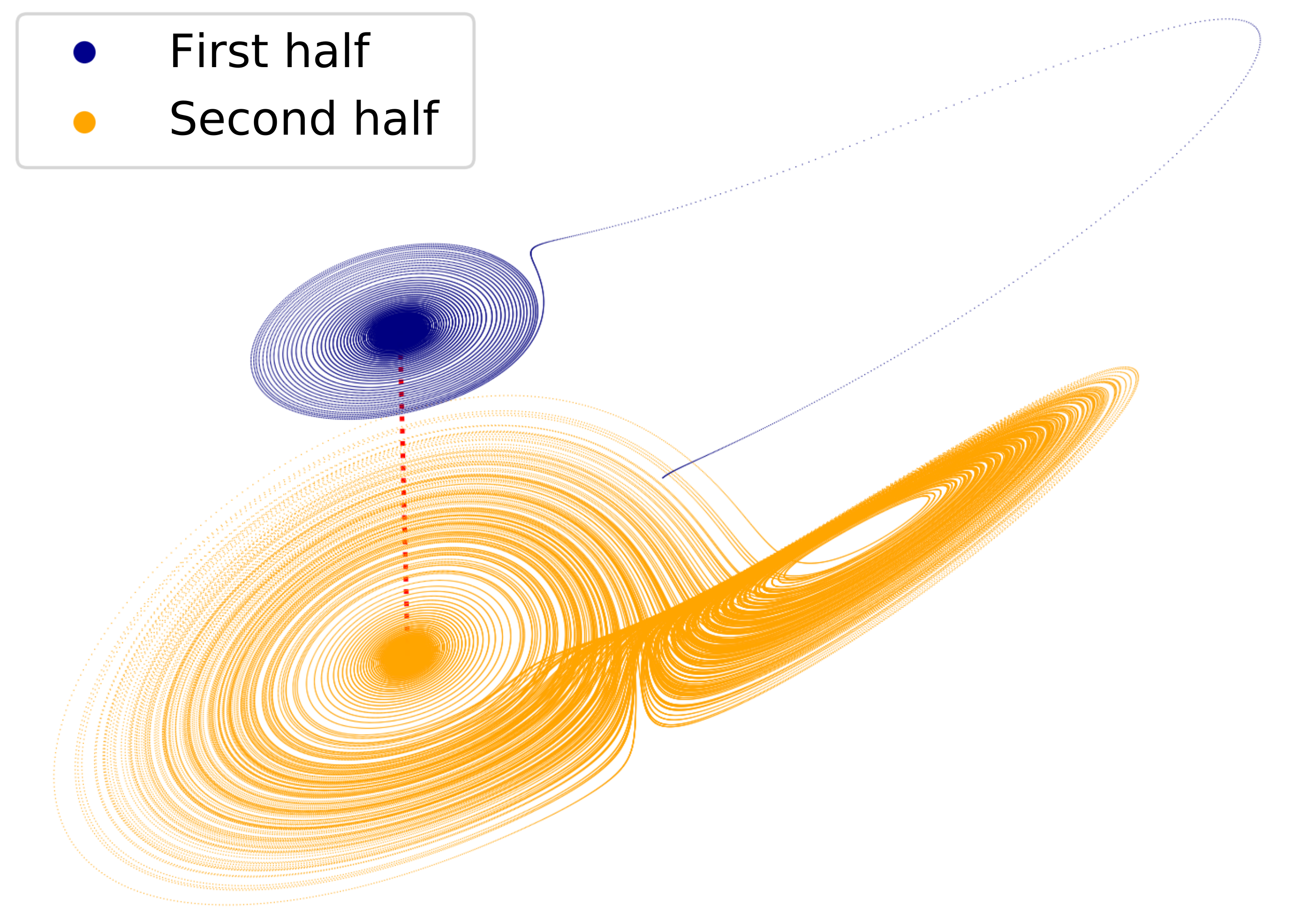}
        \caption{Representation of the first(blue) and second half (orange) of the training data. The first has parameters $\sigma = 20$, $\beta = \frac{8}{3} $, $\rho = 28 $ the second has parameters $\sigma = 10$, $\beta = \frac{8}{3} $, $\rho = 28 $.}
        \label{fig:figure2}
    \end{minipage}
    \hfill
    \begin{minipage}[h!]{0.49\textwidth}
        \centering
        \includegraphics[width=\textwidth]{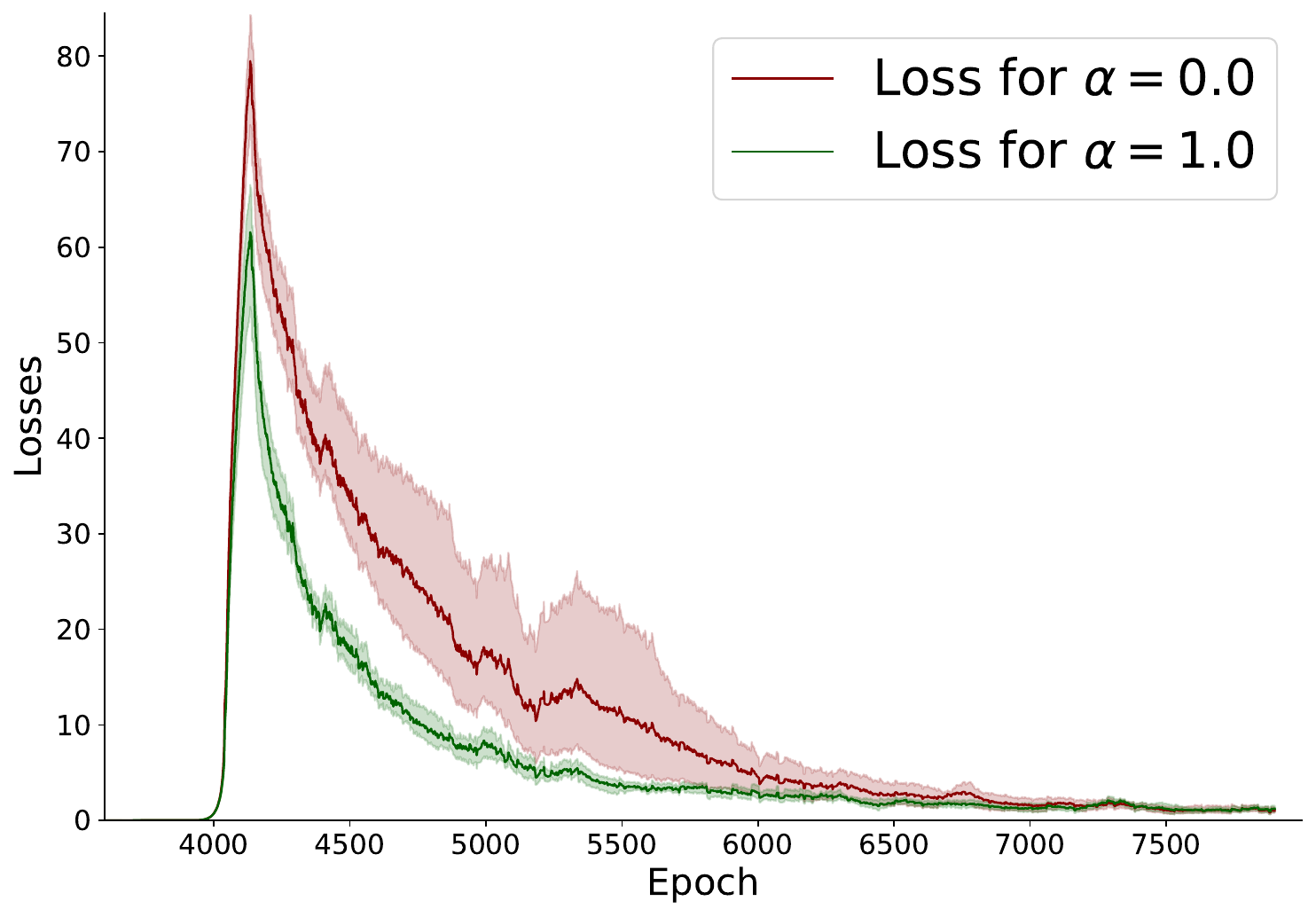}
        \caption{Example of loss behavior for the two networks. The shaded areas are the first and third quartiles, calculated from 10 training simulations for vanilla and Lyapunov networks.}
        \label{fig:figure3}
    \end{minipage}
    \label{fig:attractorsLorenz}
\end{figure}
\autoref{fig:figure2} shows the trajectories of the two dynamical systems, with the first half offset vertically for clarity. Initially, the attractor converges slowly to a limit cycle, then shifts to a double-lobed Lorenz attractor as parameters change. In \autoref{fig:figure3}, we show the loss behavior after the regime shift for $ \alpha = 1.0 $, averaged over 10 simulations. 

\paragraph{Loss‐ratio as a summary of online adaptation}
After the abrupt parameter shift in the Lorenz system, we compute the mean squared error (MSE) prediction losses for both models over the post‐shift interval \(t_0\) to \(T\), \begin{equation}
\mathcal{L}_{\mathrm{vanilla}}^{MSE} = \sum_{t=t_0}^{T}\bigl\|\mathbf{x}_{t}-\tilde{\mathbf{x}}_{t}^{\mathrm{vanilla}}\bigr\|^{2},
\end{equation}and 
\begin{equation}
    \mathcal{L}_{\mathrm{Lyap}}^{MSE}    = \sum_{t=t_0}^{T}\bigl\|\mathbf{x}_{t}-\tilde{\mathbf{x}}_{t}^{\mathrm{Lyap}}\bigr\|^{2},
    \end{equation}
We then define the \emph{loss ratio} \begin{equation}
    r = \frac{\mathcal{L}^{MSE}_{\mathrm{vanilla}}}{\mathcal{L}^{MSE}_{\mathrm{Lyap}}},
    \end{equation}
which quantifies the relative reduction in post‐shift error achieved by the Lyapunov‐regularized network compared to the vanilla baseline.

By using this ratio, we cancel out much of the run‐to‐run noise inherent in chaotic dynamics, since fluctuations that affect both models equally tend to divide out, yielding a more stable measure of comparative performance. Moreover, because our setting is truly online—where the network continuously predicts and updates without a defined training endpoint—we aggregate pure‐MSE losses over the entire post‐shift window. This ensures that \(r\) reflects sustained adaptation rather than transient recovery, faithfully representing our learning framework's perpetual predict–update loop.

This behavior echoes the concept of the Adjacent Possible: the optimal Lyapunov regularization strength (\(\alpha \approx 1.0\)) corresponds to the regime in which new dynamical possibilities can be most rapidly assimilated without excessive exploration or fixation. Empirically, we observe a mean loss ratio of \(r \approx 1.96\), indicating nearly a two‐fold reduction in MSE. By contrast, standard weight‐regularization techniques such as L1 and L2 penalties—while effective for sparsification or overfitting mitigation—yield negligible gains under abrupt regime shifts or even worsens the overall loss. This confirms that general‐purpose regularizers do not confer adaptability to non‐stationarity, highlighting the necessity of a domain‐specific Lyapunov Learning approach that steers the model’s internal dynamics toward rapid, robust adaptation.

\begin{table}[ht]
\centering
\begin{tabular}{lcc}
\toprule
Regularizer & Best Loss Ratio \(r\) & Parameter Value \\
\midrule
Dropout   & \(0.44\)           & \(P_{\text{dropout}} = 0.2\)     \\
L2        & \(0.73\)           & \(\alpha = 1\times10^{-3}\)     \\
L1        & \(1.21\)           & \(\alpha = 1\times10^{-4}\)     \\
Lyapunov  & \(1.96\)           & \(\alpha = 1.0\)                \\
\bottomrule
\end{tabular}
\caption{Best post‐shift loss ratio and corresponding hyperparameter for each regularization method.}
\label{tab:loss_ratio_comparison}
\end{table}

\begin{figure}[h!]
    \centering
    \begin{minipage}[h!]{0.49\textwidth}
        \centering
        \includegraphics[width=\textwidth]{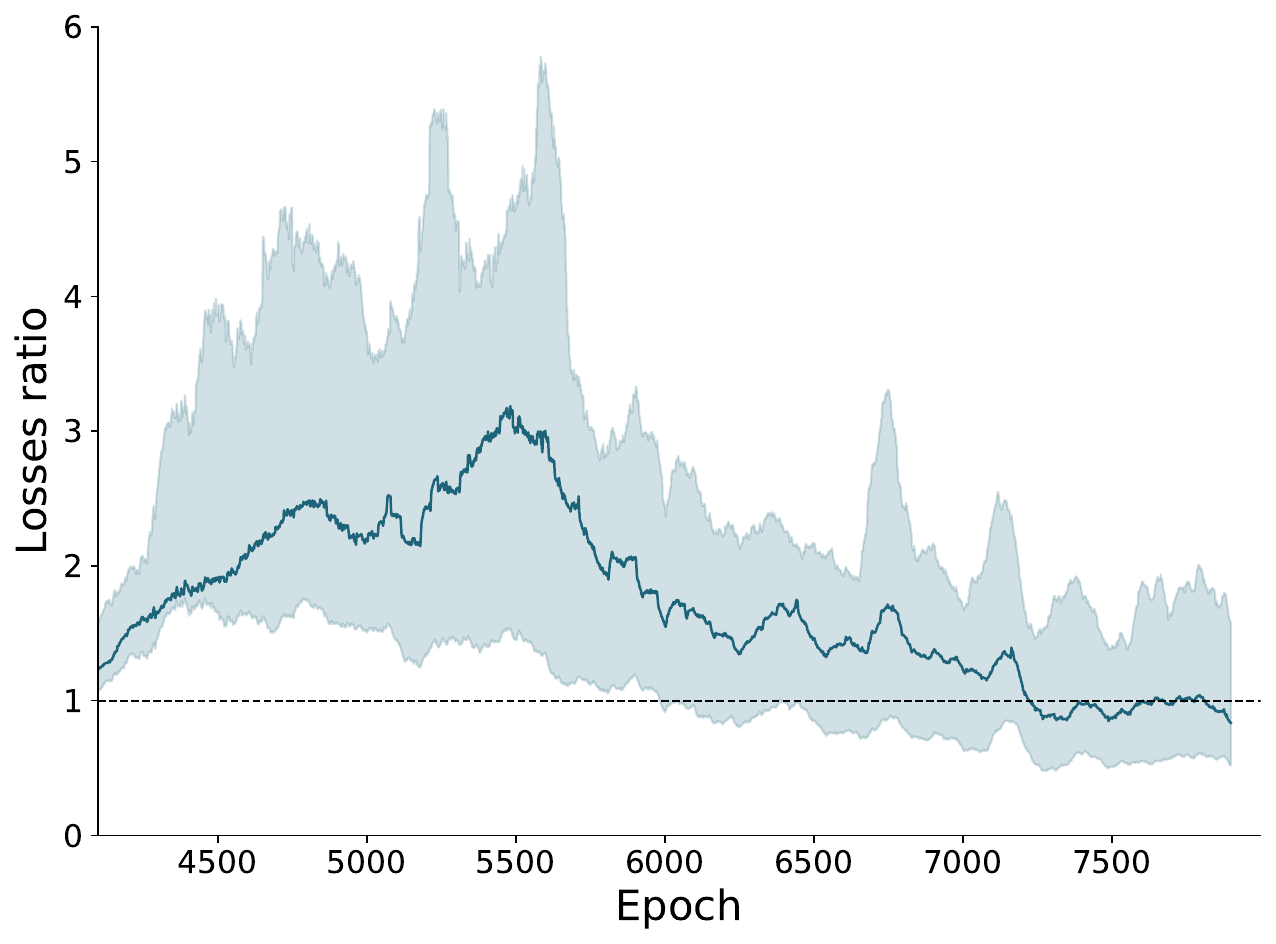}
        \caption{Losses ratio for $\alpha = 1.0$. The shaded areas are the first and third quartiles from 10 training simulations for both vanilla and Lyapunov networks.}
        \label{fig:figure4}
        \vspace{0.6em}
    \end{minipage}
    \hfill
    \begin{minipage}[h!]{0.49\textwidth}
        \centering
        \includegraphics[width=\textwidth]{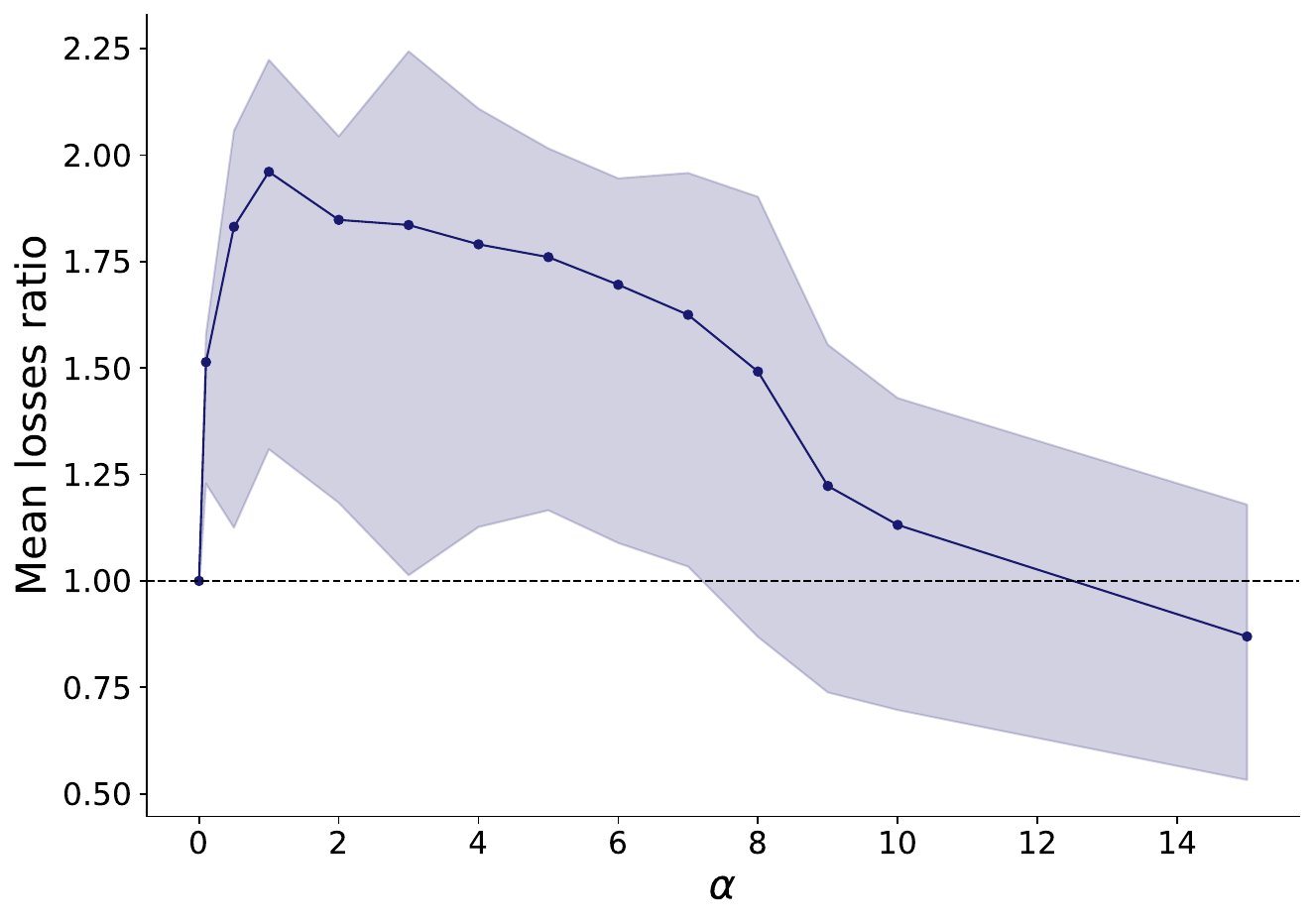}
        \caption{Mean losses ratios for different $\alpha$. Every point is the average of ten trainings. The shaded areas are the first and third quartiles of each mean ratio.}
        \label{fig:figure5}
        \vspace{-1.0em}
    \end{minipage}
    \label{fig:attractorsLorenz}
\end{figure}

\section{Conclusions}

This paper introduces a novel method for handling novelties in non-stationary real-valued time series by incorporating a chaos-driven regularizer grounded in Lyapunov exponents. While standard regularizers enhance generalization, they do not explicitly address regime transitions. Our experiments demonstrate that including Lyapunov exponents in the loss function significantly improves network performance on non-stationary data. The algorithm adapts to new patterns far more effectively than its vanilla counterpart; on Lorenz map data the post-shift mean‐squared error is nearly halved relative to the baseline, confirming the practical benefit of the proposed regularizer.

Despite its conceptual appeal, Lyapunov Learning incurs computational overhead, since assembling the state-Jacobian grows on the order of $O(d^2)$ and its QR decomposition on the order of $O(d^3)$ in the input dimension $d$, which constrains its use in deep or wide architectures. Moreover, all empirical validation has been confined to low-dimensional, noise-free chaotic systems, leaving open questions about behavior under high-dimensional, stochastic, or partially observed real-world conditions. To address these challenges, future work should pursue randomized projections or subspace-tracking techniques that capture dominant Lyapunov directions without full-Jacobian computations, thereby reducing both memory and time costs.

Finally, in many modern sequence models—whether framed as SSMs, RNNs, or linear‐attention Transformers—the core strategy for avoiding exploding or vanishing dynamics closely recalls the idea of keeping the largest Lyapunov exponent just below zero. For example, Structured State Space Models enforce spectral constraints on their recurrence matrices, stabilizing long‐range dependencies by implicitly centering dominant exponents at the edge of chaos \cite{gu2024transformers}. Likewise, recasting attention as a recurrent kernel update in \cite{katharopoulos2020fast} uses linearizations that control gradient norms analogously to orthogonal RNN initializations. Classic RNN remedies—orthonormal weight initializations, gating, and Jacobian‐norm penalties— also recall a zero‐centered Lyapunov spectrum \cite{pascanu2013vanishing}. More recent advances like Mamba apply selective subspace projections to monitor and constrain the most unstable directions in linear time \cite{gu2024mamba}, while decaying‐exponential parameterizations directly expose negative Lyapunov exponents as decay rates \cite{gu2024efficiently}.

Verifying this resemblance between Lyapunov Learning and existing stability-inducing techniques both theoretically (by relating spectral regularizers to Lyapunov exponents) and experimentally (by measuring exponent distributions across architectures) would deepen our understanding of sequence-model dynamics, unify disparate regularization strategies under a common framework, and guide the design of next-generation models that are robust yet highly adaptable.
\section*{Acknowledgments}
his work has been supported by PNRR MUR project PE0000013-FAIR.
\clearpage

\newpage
\bibliography{biblio}
\bibliographystyle{plainnat}
\comm{ \newpage
\appendix

\section*{Appendix A: Lyapunov Regularization Term – Differentiability and Implementation}

We denote the Lyapunov regularization term as \( L_{\text{Lyapunov}} \), which depends on the largest Lyapunov exponent \( \lambda_{\max} \) of the trajectories generated by the neural network. For a sequence \( \{\mathbf{x}_t\} \) recursively produced by the network \( \mathbf{x}_{t+1} = F(\mathbf{x}_t, \theta) \), we compute the Jacobian matrices
\[
J_t = \frac{\partial F(\mathbf{x}_t, \theta)}{\partial \mathbf{x}_t}
\]
at each time step. These Jacobians are then used to construct the finite-time Lyapunov exponents via the standard QR-based approximation:
\[
\Lambda = \frac{1}{T} \log \left\| \prod_{t=0}^{T-1} J_t \right\|.
\]
This method follows the approach outlined in \cite{abarbanel1997analysis}, where \( T \) is the time horizon over which the product is computed. The QR decomposition is used to numerically stabilize the product and extract the  exponents $\lambda$ .

In our implementation, all Jacobian computations are done using automatic differentiation frameworks (e.g., PyTorch's \texttt{autograd}), which makes this process fully differentiable with respect to the network weights \( \theta \). This enables the Lyapunov exponent estimate to be included in the overall loss function:
\[
L(\mathbf{x}, \hat{\mathbf{x}}) = L_{\text{data}} + \alpha \cdot |\lambda_{\max}|,
\]
and optimized via standard backpropagation. Thus, the Lyapunov Learning term acts as a gradient-compatible regularizer that pushes the system towards operating at the edge of chaos.

This formulation is crucial for enabling Lyapunov Learning in practice and is one of the distinguishing features of our approach compared to classical regularization techniques.}

\end{document}